%% file: main.tex
\documentclass[conference]{IEEEtran}

\usepackage[utf8]{inputenc}
\usepackage[T1]{fontenc}

\usepackage{amsmath}
\usepackage{amssymb}
\usepackage{graphicx}

\usepackage{booktabs}
\usepackage{tabularx}
\usepackage{array}

\usepackage{microtype}

\usepackage[numbers,sort&compress]{natbib}
\usepackage{xurl}
\usepackage[hidelinks]{hyperref}

\usepackage{tikz}
\usetikzlibrary{positioning, arrows.meta, shapes.geometric, calc}

\usepackage{enumitem}

\title{
When Compliance Data Masquerades as Evaluation:\\
Measurement Validity for Deployed AI Systems
}

\makeatletter
\newcommand{\linebreakand}{%
  \end{@IEEEauthorhalign}
  \hfill\mbox{}\par
  \mbox{}\hfill\begin{@IEEEauthorhalign}
}
\makeatother

\author{
\begin{tabular}{@{}p{0.47\textwidth}p{0.47\textwidth}@{}}

\centering
Hung-Yu Lin\\
\textit{Dept.\ of Computing Sciences}\\
\textit{Texas A\&M University--Corpus Christi}\\
Corpus Christi, TX, USA\\
hlin5@islander.tamucc.edu
&
\centering
Xingran Huang\\
\textit{Dept.\ of Computer Science and Engineering}\\
\textit{University of California, Riverside}\\
Riverside, CA, USA\\
xhuan230@ucr.edu
\tabularnewline

\noalign{\vskip 1.2em}

\centering
Qiming Guo\\
\textit{Dept.\ of Computing Sciences}\\
\textit{Texas A\&M University--Corpus Christi}\\
Corpus Christi, TX, USA\\
qguo2@islander.tamucc.edu
&
\centering
Jinwen Tang\\
\textit{EECS Department}\\
\textit{University of Missouri}\\
Columbia, MO, USA\\
jt4cc@umsystem.edu
\tabularnewline

\end{tabular}
}

\begin{document}

\maketitle


\begin{abstract}
We argue that a recurring failure in the evaluation of deployed AI systems
occurs when data collected for operational monitoring or regulatory
compliance are interpreted as if they were designed for comparative
evaluation. Automated driving provides a concrete example of this problem.
U.S. disengagement and crash-reporting regimes produce valuable operational
evidence, but differences in reporting scope, exposure, deployment domain,
event capture, and comparator construction limit the safety claims that can
be supported from these measurements alone.
We frame this issue as a measurement-validity problem in AI evaluation
rather than as a transportation-specific data limitation. We argue that
comparative claims about deployed AI systems require alignment between the
intended capability, measured outcome, exposure opportunity, deployment
domain, data-generation process, and evaluation comparator. Using
automated-driving safety evaluation as a case study, we propose an
evaluation contract that makes these assumptions explicit before operational
data are interpreted as evidence of comparative performance. The broader
implication is that data useful for monitoring deployed AI systems are not
automatically valid benchmarks for evaluating them.
\end{abstract}



\section{Introduction}

Machine-learning systems are increasingly evaluated after deployment rather
than only through static benchmarks. In high-stakes applications, deployed
systems generate operational data through logs, incident reports,
monitoring systems, and user interactions. These data are valuable because
they reflect real-world usage conditions. However, operational data are
often collected for purposes such as monitoring, compliance, debugging, or
incident investigation rather than for supporting comparative evaluation
claims.

This creates a fundamental challenge for AI evaluation. A measurement
process can accurately record the events it was designed to capture while
still failing to support a stronger inference about system capability.
For example, an incident count may be useful for detecting failures, but it
does not automatically provide a valid estimate of relative system
performance unless the outcome definition, exposure opportunity, deployment
conditions, and comparison population are appropriately aligned.

The distinction is particularly important for deployed AI systems because
their behavior is inseparable from the environments in which they operate.
Unlike controlled benchmark settings, deployment introduces variation in
users, environments, operating conditions, data availability, and failure
reporting mechanisms. Evaluation therefore depends not only on the model or
system being tested, but also on the infrastructure through which evidence
is collected.

Automated driving systems (ADS) provide a concrete example of this broader evaluation problem. The stakes extend beyond individual vehicles: automated driving interacts with traffic operations, freight logistics, infrastructure, and regulation, and its effects are system-level and policy-dependent \citep{lin2026connected}. Safety claims about deployed ADS increasingly rely on real-world observations, including disengagement reports, crash reporting systems, and retrospective mileage-based analyses. However, these sources
represent different measurement regimes rather than a single standardized evaluation benchmark.

California disengagement reports, for example, were created to describe
autonomous-vehicle testing activity and were not intended as direct
cross-company safety comparisons \citep{cadmv2026disengagement}. Similarly,
the National Highway Traffic Safety Administration (NHTSA) Standing General
Order provides valuable crash-reporting information, but its data
limitations include differences in reporting access, incomplete or
unverified information, duplicate reporting, and the absence of exposure
normalization \citep{nhtsaSGO2025,nhtsaSGOData2026}.

These limitations are not simply transportation-specific data problems.
They represent a broader issue in evaluating deployed AI systems: the
measurement process may not match the inference being made. A dataset can
be appropriate for operational monitoring while being insufficient for
ranking systems, estimating comparative performance, or supporting
deployment decisions.

In this paper, we argue that evaluation of deployed AI systems requires an
explicit connection between measurement design and the claim being made.
Using automated-driving safety evaluation as a case study, we examine how
compliance-oriented and operational data sources can be misinterpreted as
comparative benchmarks. We then identify recurring evaluation mismatches
and propose an evaluation contract that specifies the conditions required
for valid comparative claims.

The contribution of this paper is not a new safety metric, empirical
ranking of automated-driving systems, or statistical estimator. Instead,
we provide a perspective on evaluation validity: before operational data
can support claims about a deployed AI system, the evaluation protocol must
establish that the measured evidence corresponds to the capability and
comparison being claimed.

\section{Related Work}
\label{sec:related}

\subsection{Deployed AI Systems Across High-Stakes Domains}

AI systems are increasingly deployed in settings where their behavior is
observed primarily through operational data rather than controlled test
sets. In critical infrastructure, sensor-integrated AI systems monitor
physical networks in real time, such as anomaly detection for urban
underground water pipelines \citep{guo2025aquasentinel}. In healthcare,
machine-learning devices are cleared for clinical use and LLM-based systems
provide mental health support and pre-screening
\citep{wu2021medical,guo2024soullmate,tang2024prescreening}, while clinical
deployments face well-documented dataset shift between development and
practice \citep{finlayson2021clinician}. In education, conversational
tutoring systems adapt to individual learners during live interaction
\citep{tang2025rpkt}. Across these domains, the software-engineering
literature has long observed that production ML systems accumulate
evaluation and monitoring debt that static benchmarks do not capture
\citep{sculley2015hidden,breck2017mltest,amodei2016concrete}. These deployed
systems all generate operational data as a byproduct of their function,
which motivates the central question of this paper: under what conditions
such data can support evaluative claims.

\subsection{AI Evaluation Methodology and Infrastructure}

A substantial literature examines the reliability of AI evaluation itself.
Reproduction studies show that benchmark conclusions can shift on newly
collected test sets \citep{recht2019imagenet}, and benchmark reuse,
leakage, and contamination further weaken evaluation claims
\citep{kapoor2023leakage,bowman2021fix,paullada2021data}. Responses include
dynamic benchmark construction \citep{kiela2021dynabench}, holistic
multi-metric evaluation \citep{liang2023helm}, cost-aware benchmarking of
LLM pipelines \citep{sun2024cebench}, and analyses of the reliability of
LLM-based judges \citep{zheng2023judging}. A complementary line of work
addresses evaluation transparency and governance through documentation
artifacts such as datasheets and model cards
\citep{gebru2021datasheets,mitchell2019modelcards} and through internal
auditing frameworks \citep{raji2020closing}. For deployed systems, the
integrity of the data-generation process itself becomes an attack surface:
adversaries can inject content into model outputs and agent pipelines
\citep{guo2025aea}, and privacy regulations may require removing data from
trained models after deployment \citep{guo2025stgunlearning}, both of which
alter the evidence available for evaluation. This literature focuses
largely on benchmark-style evaluation; comparatively little work asks when
operational and compliance data can serve as evaluation evidence, which is
the gap this paper addresses.

\subsection{Measurement Validity and Automated-Driving Safety Evaluation}

Our framing draws on measurement theory. Construct validity originates in
psychometrics \citep{cronbach1955construct,messick1995validity} and has
been applied to machine learning through measurement modeling of fairness
constructs \citep{jacobs2021measurement}, critiques of benchmark validity
\citep{raji2021benchmark,miller2022validity,bean2025construct}, and recent
position work framing generative AI evaluation as a social-science
measurement problem \citep{wallach2025position}.

In automated driving, prior work has quantified how many miles of
operation are required to statistically demonstrate safety
\citep{kalra2016driving}, proposed frameworks for measuring automated
vehicle safety \citep{fraadeblanar2018measuring}, and analyzed the
limitations of California disengagement data as a comparative measure
\citep{favaro2017examining,banerjee2018hands}. Operator-published analyses
of deployed mileage \citep{schwall2020waymo,kusano2024waymo} and the RAVE
checklist for retrospective safety studies \citep{scanlon2025rave} address
exposure and benchmark construction, while safety-case standards and
testing-validation analyses structure single-system safety argumentation
\citep{ul4600,koopman2016challenges}, with operational scope formalized
through ODD definitions \citep{sae2021j3016}.

Our work differs from these efforts in scope and target. Safety cases and
UL~4600 argue the acceptability of a single system; RAVE provides
methodological recommendations specific to retrospective ADS crash-rate
studies. The evaluation contract proposed here instead specifies the
validity conditions under which operational and compliance data from any
deployed AI system can support \emph{comparative} claims, treating ADS
reporting regimes as one instance of a general measurement problem.

\section{Measurement Regimes That Look Comparable but Are Not}
\label{sec:regimes}

Automated-driving safety evaluation provides a useful case study because
multiple public data sources appear to describe similar quantities:
events, mileage, vehicles, and operational outcomes. However, these sources
were created for different purposes and therefore support different types
of inference.

The central issue is not whether any individual dataset is useful. Each
measurement regime provides valuable information for its intended purpose.
The issue is whether evidence generated for monitoring, compliance, or
incident reporting can support a stronger comparative claim without
additional assumptions.

Table~\ref{tab:measurement_regimes} summarizes the distinction between
these regimes.

\begin{table}[t]
\centering
\small
\setlength{\tabcolsep}{4pt}
\caption{
Measurement regimes in automated-driving evaluation and the comparative claims they can support.
}
\label{tab:measurement_regimes}
\begin{tabular}{p{0.20\columnwidth}
                p{0.30\columnwidth}
                p{0.34\columnwidth}}
\toprule
\textbf{Regime}
&
\textbf{Measures}
&
\textbf{Cannot establish alone}
\\
\midrule
California DMV
&
Testing mileage and disengagement events
&
Comparable system safety across operators
\\
\addlinespace
NHTSA SGO
&
Reported crash incidents
&
Exposure-normalized risk comparison
\\
\addlinespace
Retrospective studies
&
Outcome rates using exposure and benchmarks
&
General conclusions beyond matched domains
\\
\bottomrule
\end{tabular}
\end{table}

\subsection{California Disengagement Reporting}

California's autonomous-vehicle disengagement reports are one of the most
visible sources of public ADS operational data. Under the historical
reporting framework, manufacturers testing autonomous vehicles on public
roads reported disengagement events and testing mileage to the California
Department of Motor Vehicles (DMV).

These reports provide useful information about system behavior during
testing. However, they were not designed as a benchmark for ranking
autonomous systems. The DMV describes the reports as records of individual
permit-holder activity rather than direct comparative evaluations across
companies \citep{cadmv2026disengagement}.

The limitation is therefore not simply statistical. Different developers
may select different vehicles, routes, environments, and testing
strategies, and prior analyses of these reports document substantial
heterogeneity in reporting practices and operating conditions across
manufacturers \citep{favaro2017examining,banerjee2018hands}. The observed
disengagement rate reflects both system behavior and the conditions under
which the system was evaluated.

The reporting scope also affects interpretation. Historical reports did not
capture all development activity, including certain activities outside
California, private-road testing, and simulation-based development.
Furthermore, testing and deployment operations have been subject to
different reporting requirements \citep{cadmv2026disengagement}.

The measurement protocol itself has also changed over time. California
adopted updated autonomous-vehicle regulations that expand safety-related
reporting requirements beyond the previous disengagement framework
\citep{cadmv2026regulations}. Comparisons across reporting periods must
therefore account for changes in what is measured and how observations are
collected.

\subsection{NHTSA Standing General Order Reporting}

The National Highway Traffic Safety Administration (NHTSA) Standing General
Order (SGO) provides another important source of ADS safety information.
The SGO requires specified manufacturers and operators to report qualifying
crashes involving automated-driving systems and Level 2 driver-assistance
systems \citep{nhtsaSGO2025}.

Unlike disengagement reports, the SGO focuses on crash events rather than
testing interventions. This makes it valuable for incident surveillance.
However, incident surveillance and comparative evaluation are different
measurement objectives.

NHTSA documentation identifies several limitations affecting direct
comparison. Reported information may initially be incomplete or unverified,
entities may differ in access to crash information, and multiple reports
may correspond to the same incident \citep{nhtsaSGOData2026}.

More importantly, incident counts do not directly represent comparative
risk. A system operating more miles, serving more users, or encountering
more complex environments has more opportunities to produce reportable
events. Without an appropriate exposure model, observed counts combine
system behavior with deployment scale and operating conditions.

The SGO therefore provides important monitoring capability, but additional
evaluation design is required before incident observations can support
comparative safety claims.

\subsection{Exposure-Adjusted Retrospective Evaluation}

Exposure-adjusted studies represent a different measurement regime because
they attempt to connect observed outcomes with opportunities for those
outcomes to occur.

Kusano et al. analyzed 7.14 million rider-only miles of Waymo automated
driving and compared observed crash outcomes with constructed human-driver
benchmarks \citep{kusano2024waymo}. By introducing an exposure denominator,
this approach addresses a central limitation of raw event counts.

However, exposure normalization alone does not resolve all evaluation
challenges. The validity of the comparison depends on benchmark
construction, operating conditions, geography, and Operational Design
Domain (ODD) \citep{sae2021j3016}.

The RAVE checklist emphasizes that retrospective ADS safety studies should
align crash outcomes and exposure measures while accounting for geography,
road type, operating conditions, and ODD
\citep{scanlon2025rave}. These requirements illustrate a broader principle:
additional data improve evaluation only when they reduce uncertainty about
the relationship between observation and claim.

The three regimes therefore represent different levels of evaluative
strength: regulatory monitoring provides visibility into operational
events; incident reporting provides safety-event surveillance; and
exposure-adjusted studies provide stronger comparative evidence.

The evaluation challenge occurs when evidence collected for one purpose is
interpreted as supporting a stronger claim than the measurement process was
designed to answer.

\section{Failure Modes in Deployed AI Evaluation}
\label{sec:failures}

The measurement regimes described in Section~\ref{sec:regimes} are not
failures because they lack data. They fail when observations generated for
one purpose are interpreted as evidence for a different claim.

This distinction reveals recurring failure modes in the evaluation of
deployed AI systems. A metric can be measured consistently and reported
accurately while still failing to identify the capability that an evaluator
intends to compare. The problem is therefore not only data availability,
but whether the measurement process supports the intended inference.

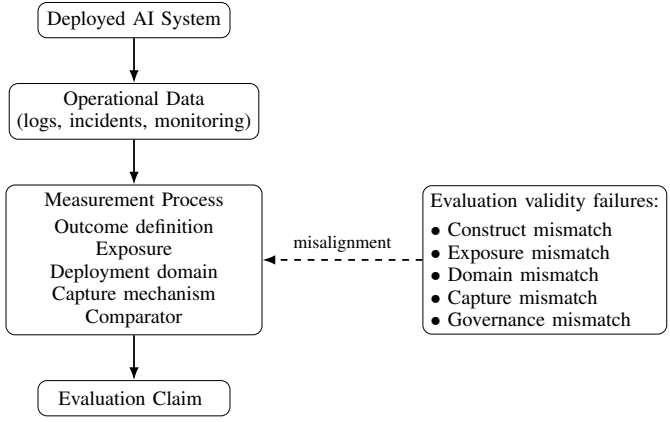
\begin{figure}[t]
    \centering
    \resizebox{\columnwidth}{!}{\input{figures/evaluation_failure.tex}}
    \caption{
    Evaluation validity depends on whether the measurement process preserves
    the assumptions required for the intended inference. Operational data can
    be accurately collected while still being insufficient to support a
    comparative claim.
    }
    \label{fig:evaluation_failure}
\end{figure}

Figure~\ref{fig:evaluation_failure} summarizes this relationship. The
failure occurs when the path from operational observation to evaluation
claim contains assumptions that are not explicitly defined or validated.

\subsection{Construct Mismatch: The Measurement Is Not the Capability}

The first failure mode occurs when an observed quantity is interpreted as a
direct measure of a broader capability.

In automated driving, disengagements and reported crashes are observable
events. However, comparative safety claims concern a larger construct:
whether an AI-enabled system can operate safely under a specified set of
conditions.

A reporting system may therefore measure its target event accurately while
still failing to measure the capability implied by the final claim. A
lower event count does not automatically indicate superior system safety
unless the relationship between the event and the intended capability has
been established.

This issue is recognized throughout machine-learning evaluation.
Benchmark metrics are informative only when the measured task corresponds
to the capability that the benchmark is intended to represent. Measurement
modeling makes this correspondence explicit by treating the link between an
observable proxy and an unobservable construct as an assumption requiring
validation \citep{jacobs2021measurement}. Recent work on construct validity
and benchmark validity likewise emphasizes that evaluation claims depend on
whether the measured phenomenon, benchmark design, and interpretation
remain aligned with the capability being assessed
\citep{bean2025construct,miller2022validity}.

The broader implication is that deployed AI evaluation must distinguish
between monitoring signals and validated measurements of system capability.

\subsection{Exposure Mismatch: The Denominator Defines the Question}

The second failure mode occurs when event counts are interpreted without a
definition of exposure.

An observed event is generated through an interaction between system
behavior and opportunity. A system operating more miles, serving more
users, or encountering more complex conditions naturally has more
opportunities to generate observable failures.

Therefore, the numerator alone does not define comparative performance.
The evaluation must specify the population over which the event rate is
interpreted.

This issue appears directly in automated-driving safety evaluation. Crash
counts or disengagement counts represent observed events, not risk
estimates, unless the corresponding exposure opportunity is defined.
Moreover, because severe outcomes are rare, statistically demonstrating
safety differences requires exposure at a scale of hundreds of millions of
miles, far beyond what most reporting regimes capture
\citep{kalra2016driving}.

The same principle applies to other deployed AI systems. A model deployed
to more users or more diverse environments may accumulate more failures
without having lower underlying reliability. Exposure is therefore part of
the evaluation target rather than a secondary correction added after the
fact.

\subsection{Domain Mismatch: Deployment Conditions Define Evaluation Scope}

The third failure mode concerns differences in deployment distribution.

Machine-learning evaluation commonly treats distribution shift as a
challenge to generalization. However, for deployed AI systems, the
deployment distribution defines the population to which the evaluation
claim applies.

The WILDS benchmark demonstrates that real-world ML systems frequently
encounter deployment distributions that differ from development
conditions, making evaluation under realistic shifts a central challenge
\citep{koh2021wilds}.

Automated driving illustrates this issue through Operational Design Domains
(ODDs). A system operating under specific geographic, weather, roadway,
and traffic conditions is evaluated over a particular distribution of
experiences. Expanding the conclusion beyond that domain requires
additional evidence.

This principle extends beyond autonomous driving. For any deployed AI
system, the question is not only whether the system performs well, but over
which users, environments, and conditions that performance has been
measured.

Domain specification is therefore not a secondary qualification added
after evaluation. It is part of the claim itself.

\subsection{Capture and Governance Mismatch: Who Defines the Evidence}

The final failure mode concerns the process through which evaluation data
are generated.

Operational data are not direct observations of system capability. They
are produced through measurement choices: which events are logged, how
events are defined, which outcomes are available, and how missing
information is handled.

These choices can influence the evidence available for evaluation.
Differences in telemetry access, reporting requirements, and data
collection procedures may create apparent performance differences that
partly reflect the measurement process itself.

A related issue concerns the relationship between evaluator access and
independence. In many deployed AI systems, the developer or operator
possesses the most complete exposure information, telemetry, and failure
context required for meaningful evaluation. However, access to these
resources may create challenges for independent assessment because the
organization providing the evidence may also have an interest in the
evaluation outcome.

This does not imply that operator-generated evaluation is invalid. In many
cases, operator access is necessary to construct a technically meaningful
evaluation. Instead, the challenge is ensuring that evaluation conclusions
remain interpretable through transparency regarding data access, measurement
definitions, benchmark construction, and analytical assumptions.

The Waymo retrospective safety analysis illustrates this access--independence
tension. Its access to system-level mileage enabled an exposure-adjusted
comparison that public incident databases cannot provide
\citep{schwall2020waymo,kusano2024waymo}, while operator involvement
highlights the importance of transparent methodology and independent
replication where feasible.

\subsection{The Common Structure of Evaluation Failure}

Although the examples above arise from automated driving, they represent a
broader pattern in deployed AI evaluation.

The failure is not caused by the absence of data. It occurs when the
measurement process does not preserve the assumptions required to connect
an observation to a claim.

A deployed AI evaluation can therefore fail through several pathways:
the measured event does not represent the claimed capability; the exposure
opportunity is undefined or incomparable; the deployment domain differs
from the claimed scope; or the data-generation process introduces
uncontrolled differences.

These mismatches motivate the evaluation contract proposed in the next
section: not as a new benchmark, but as a specification of the conditions
under which deployed-system evidence can support comparative claims.

\section{An Evaluation Contract for Deployed AI Systems}
\label{sec:contract}

The evaluation failures described in Section~\ref{sec:failures} suggest
that deployed AI evaluation requires more than collecting additional
observations. The central requirement is an explicit relationship between
the evidence being measured and the claim being made.

We refer to this relationship as an \textit{evaluation contract}. An
evaluation contract is not a new metric or a universal benchmark. Instead,
it is a specification of the assumptions required for interpreting
operational data as evidence for a particular evaluation claim.

The purpose of an evaluation contract is therefore not to eliminate
uncertainty. Rather, it makes the connection between measurement and
inference explicit so that the validity of a conclusion can be examined.

Table~\ref{tab:evaluation_contract} summarizes the relationship between
common evaluation failures and the corresponding requirements of an
evaluation contract.

\begin{table}[t]
\centering
\small
\setlength{\tabcolsep}{4pt}
\caption{
Mapping between evaluation failure modes and corresponding contract requirements.
}
\label{tab:evaluation_contract}
\begin{tabular}{p{0.28\columnwidth}
                p{0.58\columnwidth}}
\toprule
\textbf{Failure mode} &
\textbf{Evaluation contract requirement} \\
\midrule
Construct mismatch &
Define the capability being evaluated and ensure that the measured outcome
represents the intended construct. \\
\addlinespace
Exposure mismatch &
Specify the opportunity over which performance is measured, including an
appropriate denominator or interaction population. \\
\addlinespace
Domain mismatch &
Define the deployment distribution and the scope over which the evaluation
claim applies. \\
\addlinespace
Capture mismatch &
Document how observations are generated, classified, and made available
for analysis. \\
\addlinespace
Governance mismatch &
Specify data access, comparator construction, transparency requirements,
and evaluation independence. \\
\bottomrule
\end{tabular}
\end{table}

\subsection{Evaluation Must Begin With the Claim}

A recurring mistake in deployed AI evaluation is beginning with available
data rather than with the decision that the evaluation is intended to
support.

Operational systems often produce large quantities of logs, incidents,
and performance statistics. However, the availability of a measurement
does not determine its validity for every possible claim.

An evaluation contract reverses this process. The evaluator first defines
the capability or decision of interest, then determines what evidence is
required to support that claim. This prevents operational metrics from
being interpreted beyond the construct they represent.

For example, a deployment-monitoring metric may be highly valuable for
detecting failures while still being unsuitable for ranking two deployed AI
systems. The distinction is not whether the metric is useful, but whether
the metric answers the question being asked.

\subsection{The Access--Independence Tension}

A central challenge for deployed AI evaluation is that the most informative
data are often controlled by the system operator.

Exposure information, detailed logs, environmental conditions, and failure
contexts are frequently available only to the organization developing or
operating the system. Without this information, external evaluators may be
unable to perform meaningful comparisons.

However, the same access that enables stronger evaluation can create
questions about independence. An operator may define event categories,
select comparison baselines, or determine which internal measurements are
available for analysis.

This creates a fundamental tension: the evaluator with the strongest access
to evidence may not be the evaluator with the greatest independence from
the outcome.

The solution is not to exclude operator involvement. For many deployed AI
systems, such exclusion would make rigorous evaluation impossible. Instead,
evaluation contracts should require transparency regarding data collection,
measurement definitions, analysis procedures, and limitations.

Independent replication, external auditing \citep{raji2020closing}, and
controlled data access mechanisms can reduce this tension by allowing
conclusions to be examined without requiring all evaluation capabilities to
exist outside the system operator.

This challenge is particularly relevant for high-stakes AI systems, where
evaluation results may influence deployment decisions, regulation, and
public trust.

\subsection{Evaluation Infrastructure Should Be Designed Before Deployment}

The evaluation contract also changes how AI system development should
approach deployment.

Evaluation is often treated as an activity performed after a system has
already been built. However, once a system is deployed, important
information about exposure, failures, and operating conditions may no
longer be recoverable if it was not collected from the beginning.

Therefore, evaluation readiness should be considered part of system design.
Developers and regulators should consider what outcomes future evaluations
may need to measure, what exposure information is required, how deployment
domains should be documented, and how independent analysis can be enabled.

This does not imply that every deployed AI system requires identical
evaluation infrastructure. Different applications require different
measurements. The broader principle is that evidence generation should be
designed together with the claims that the system may eventually need to
support.

\subsection{From Benchmarks to Evidence Contracts}

Traditional benchmarks remain essential for measuring progress under
controlled conditions. However, deployed AI systems introduce additional
evaluation requirements because their behavior depends on changing users,
environments, and operating conditions.

The evaluation contract perspective does not replace benchmarks. It extends
evaluation beyond fixed datasets by requiring explicit assumptions about
how evidence connects to real-world claims.

For deployed AI systems, the central question is therefore not ``What
score did the system achieve?'' but ``What claim does this measurement
support, under which conditions, and with what assumptions?''

This shift from metric-centered evaluation to claim-centered evaluation is
the main implication of the evaluation contract.

\section{Implications for AI Evaluation Infrastructure}

Automated driving provides a concrete example of a broader challenge in
deployed AI evaluation. The central issue is not specific to transportation,
but arises whenever operational data collected from real-world systems are
used to support claims about system capability.

Traditional machine-learning benchmarks provide value because the evaluation
population, task definition, and measurement procedure are explicitly
specified. Deployed systems introduce additional uncertainty because users,
environments, operating conditions, and failure mechanisms evolve after
deployment.

The implication is that evaluation infrastructure should be considered part
of the AI system itself. Data collection policies, logging mechanisms,
failure definitions, and reporting procedures determine what conclusions
can later be drawn about system performance. If these elements are not
designed with evaluation objectives in mind, important questions may become
impossible to answer after deployment.

This perspective does not suggest replacing benchmarks with operational
monitoring. Benchmarks and monitoring serve different purposes. Controlled
benchmarks measure performance under defined conditions, while operational
data reveal system behavior in real environments. Reliable evaluation of
deployed AI systems requires understanding how these sources of evidence
relate to the claims being made.

The same evaluation challenge can appear in other deployed AI domains,
including medical systems \citep{wu2021medical,finlayson2021clinician},
robotics, and autonomous agents. In each case, the relevant questions
remain the same: what capability is being measured, over which population
does the claim apply, how are observations generated, and whether the
evidence supports the intended decision.

The broader lesson is that evaluation validity depends not only on the
amount of data collected, but on whether the measurement infrastructure was
designed to support the claims made from it.

\section{Limitations and Scope}

This paper presents a position on evaluation validity rather than an
empirical evaluation of automated-driving systems. We do not estimate
safety rates, compare autonomous-driving systems, or propose a new
statistical estimator. Instead, we analyze how measurement processes
influence the types of claims that deployed AI system data can support.

The analysis is based on automated driving as a high-stakes case study.
Although the evaluation challenges identified here are relevant to other
deployed AI systems, the extent to which the proposed evaluation contract
transfers across domains requires further investigation.

This paper also does not argue that operational monitoring data are
insufficient or unnecessary. Regulatory reports, telemetry, and deployment
logs remain valuable for incident detection, debugging, and system
improvement. The narrower claim is that evidence collected for one purpose
should not automatically be interpreted as supporting a different
evaluation objective without validating the connection between measurement
and inference.

Finally, the evaluation contract proposed here is conceptual. Future work
should investigate how these principles can be operationalized through
specific evaluation protocols, auditing processes, and deployment
infrastructure for different classes of AI systems.


\clearpage
\bibliographystyle{IEEEtranN}
\bibliography{references}

\end{document}

%% file: figures/evaluation_failure.tex
\begin{tikzpicture}[
    node distance=0.7cm,
    box/.style={
        draw,
        rounded corners,
        align=center,
        minimum width=3.0cm,
        minimum height=0.55cm,
        font=\small
    },
    largebox/.style={
        draw,
        rounded corners,
        align=center,
        minimum width=4.0cm,
        minimum height=1.1cm,
        font=\small
    },
    fail/.style={
        draw,
        rounded corners,
        align=left,
        minimum width=3.0cm,
        font=\small
    },
    arrow/.style={
        -{Latex[length=2mm]},
        thick
    }
]
\node[box] (system)
{Deployed AI System};
\node[box, below=of system] (data)
{Operational Data\\
(logs, incidents, monitoring)};
\node[largebox, below=of data] (measurement)
{
Measurement Process\\[2pt]
Outcome definition\\
Exposure\\
Deployment domain\\
Capture mechanism\\
Comparator
};
\node[box, below=of measurement] (claim)
{
Evaluation Claim
};
\draw[arrow] (system) -- (data);
\draw[arrow] (data) -- (measurement);
\draw[arrow] (measurement) -- (claim);
\node[fail, right=2.5cm of measurement] (fail)
{
Evaluation validity failures:\\[3pt]
$\bullet$ Construct mismatch\\
$\bullet$ Exposure mismatch\\
$\bullet$ Domain mismatch\\
$\bullet$ Capture mismatch\\
$\bullet$ Governance mismatch
};
\draw[arrow, dashed]
(fail.west) -- 
node[above,font=\footnotesize]{misalignment}
(measurement.east);
\end{tikzpicture}